\documentclass[conference]{IEEEtran}
\IEEEoverridecommandlockouts

\usepackage{cite}
\usepackage{amsmath}
\usepackage{amssymb}
\usepackage{amsfonts}
\usepackage{algorithmic}
\usepackage{graphicx}
\usepackage{textcomp}
\usepackage[dvipsnames]{xcolor}
\usepackage{hyperref}
\usepackage{arydshln}
\usepackage{booktabs}
\usepackage{cleveref}
\usepackage[labelfont=bf, labelsep=colon]{caption}

\AtEndPreamble{
    \crefname{figure}{Fig.}{Figs.}
    \Crefname{figure}{Fig.}{Figs.}
    \crefname{table}{TABLE}{TABLES}
    \Crefname{table}{TABLE}{TABLES}
}

\definecolor{custompink}{rgb}{0.99, 0.03, 0.51}
\def\BibTeX{{
    \rm B\kern-.05em{\sc i\kern-.025em b}\kern-.08em
    T\kern-.1667em\lower.7ex\hbox{E}\kern-.125emX
}}

\usepackage{amsmath,amsfonts,bm}

\def\eqref#1{equation~\ref{#1}}

\def\1{\bm{1}}

\def\vc{{\bm{c}}}
\def\vd{{\bm{d}}}

\def\vh{{\bm{h}}}

\def\vs{{\bm{s}}}

\def\vv{{\bm{v}}}

\def\vx{{\bm{x}}}

\def\mC{{\bm{C}}}

\def\mH{{\bm{H}}}
\def\mI{{\bm{I}}}

\def\mY{{\bm{Y}}}

\DeclareMathAlphabet{\mathsfit}{\encodingdefault}{\sfdefault}{m}{sl}
\SetMathAlphabet{\mathsfit}{bold}{\encodingdefault}{\sfdefault}{bx}{n}

\def\gL{{\mathcal{L}}}

\DeclareMathOperator*{\argmin}{arg\,min}

\begin{document}

\title{
Face-StyleSpeech: Enhancing Zero-shot Speech Synthesis from Face Images with Improved Face-to-Speech Mapping
}

\author{
\IEEEauthorblockN{
    \textbf{Minki Kang}$^{*}$\thanks{$^*$: Equal Contribution}
}
\IEEEauthorblockA{
    KAIST \\
    \texttt{zzxc1133@kaist.ac.kr}
}
\and
\IEEEauthorblockN{
    \textbf{Wooseok Han}$^{*}$
}
\IEEEauthorblockA{
    AITRICS \\ \texttt{hwrg@aitrics.com}
}
\and
\IEEEauthorblockN{
    \textbf{Eunho Yang}
}
\IEEEauthorblockA{
    KAIST, AITRICS \\ \texttt{eunhoy@kaist.ac.kr}
}
}

\maketitle

\begin{abstract}
Generating speech from a face image is crucial for developing virtual humans capable of interacting using their unique voices, without relying on pre-recorded human speech.
In this paper, we propose Face-StyleSpeech, a zero-shot Text-To-Speech (TTS) synthesis model that generates natural speech conditioned on a face image rather than reference speech.
We hypothesize that learning entire prosodic features from a face image poses a significant challenge.
To address this, our TTS model incorporates both face and prosody encoders. The prosody encoder is specifically designed to model speech style characteristics that are not fully captured by the face image, allowing the face encoder to focus on extracting speaker-specific features such as timbre.
Experimental results demonstrate that Face-StyleSpeech effectively generates more natural speech from a face image than baselines, even for unseen faces.
Samples are available on our demo page.\footnote{\href{https://face-stylespeech.github.io}{https://face-stylespeech.github.io}}
\end{abstract}

\section{Introduction}

The zero-shot TTS model~\cite{NeuralVoiceCloning} is designed to synthesize speech that closely resembles the target speaker’s voice using reference speech, without needing any fine-tuning.
Recent works~\cite{StyleSpeech, YourTTS, Guided-TTS-2, GradStyleSpeech} have utilized a speech encoder to embed reference speech into a latent speech vector, then reconstructed speech based on this vector and the transcript during training. This speech vector captures both prosody and timbre, essential for synthesizing natural-sounding speech from any speaker.

While traditional TTS models rely on reference speech to replicate a target speaker’s voice, generating speech from a face image provides a promising direction for creating novel voices without needing reference speech.
Previous works~\cite{Face2Speech, FR-PSS, ImaginaryVoice} have validated this approach by mapping or replacing the speech vector with a latent face vector derived from a face image.
However, creating natural-sounding speech from just a face image remains a significant challenge, as the face alone cannot fully capture all the complex characteristics of speech.
While facial features like gender, age, and race provide some clues to speaker identity and broader elements like timbre and pitch range can be inferred, they fail to capture the more intricate prosodic features such as precise pitch, stress, and intonation.
Therefore, modeling these nuanced \textit{speech style} is crucial for accurate speech synthesis from face images.

\begin{figure}
    \centering
    \includegraphics[width=0.88\linewidth] {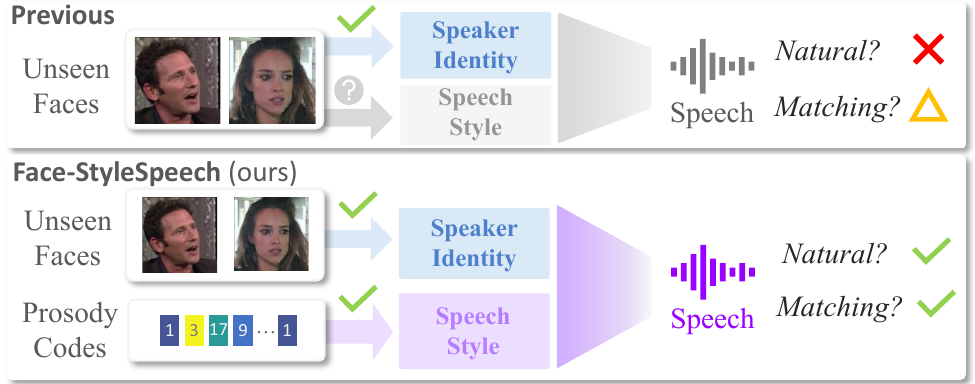}
    \vspace{-0.07in}
    \caption{\textbf{Concept.} 
    Previous methods model both speaker-identity and speech style from the face image, which is highly challenging. Instead, we model the speech style with the prosody codes to improve the stability of face-to-voice mapping only on speaker-related features.}
    \vspace{-0.2in}
    \label{fig:concept}
\end{figure}

In this work, we introduce \textbf{Face-StyleSpeech}, a zero-shot TTS model that generates natural speech from facial images. By focusing on distinct speech style features, we enhance the mapping from face-to-speech, enabling more accurate voice generation. 
This is accomplished by incorporating a prosody encoder\cite{megatts}, which zero-shot TTS model~\cite{GradStyleSpeech} speech style features into prosody codes, allowing the face encoder to specialize in extracting speaker identity information.
The face encoder is trained to align the face vector with the speech vector, ensuring that the voice generated from a face image accurately reflects the speaker’s identity.

In experiments, we use the LibriTTS-R dataset~\cite{LibriTTS-R} and the VoxCeleb dataset~\cite{VoxCeleb} to train the TTS model and the face encoder, respectively.
Our evaluation, using unseen face images demonstrate that Face-StyleSpeech consistently generates speech that is more natural compared to baselines, significantly improving the alignment between voice and facial cues.

\begin{figure*}
    \centering
    \includegraphics[width=1.0\linewidth]{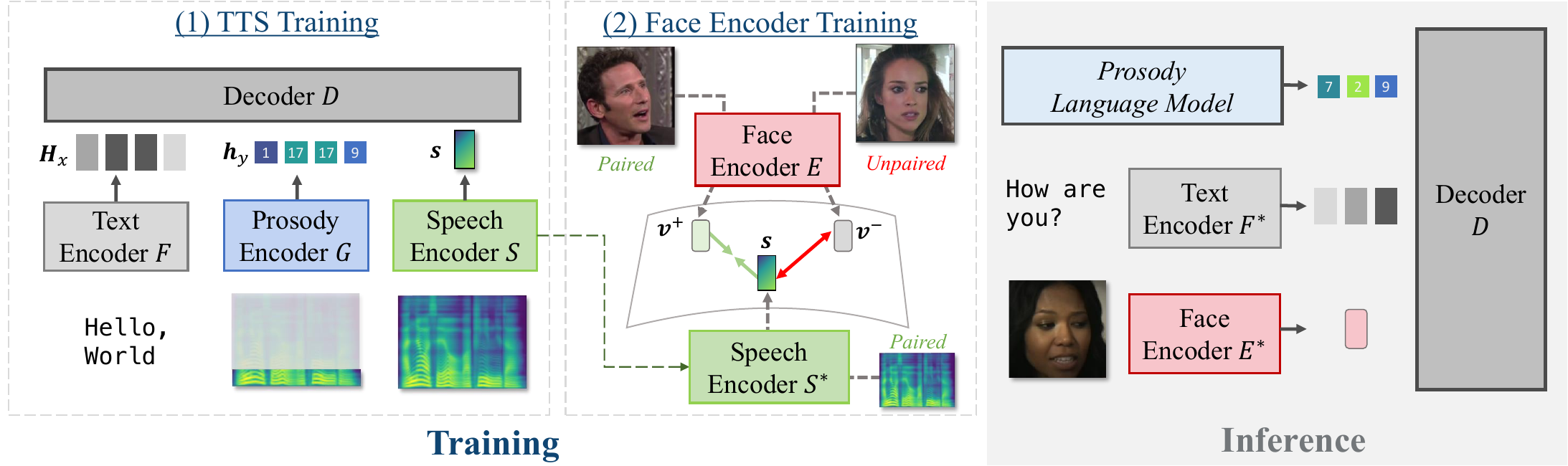}
    \vspace{-0.2in}
    \caption{The overview of Face-StyleSpeech. (1) The TTS model generates speech given the text embedding, prosody codes, and the speech vector. (2) We train the face encoder to generate a face vector corresponding to the paired speech vector. (3) In inference, we use the face vector and prosody codes from the prosody language model.}
    \vspace{-0.25in}
    \label{fig:method}
\end{figure*}

Our contributions are as follows:
\begin{itemize}
    \item We present \emph{Face-StyleSpeech}, a zero-shot TTS model that synthesizes natural speech from a face image.
    
    \item We introduce the use of prosody codes to represent distinct speech style features, such as prosody and timbre, improving the mapping between face images and speech in the latent vector space.
    
    \item In experiments, we empirically show that the proposed method significantly improves the naturalness of the speech generated from unseen face images.
\end{itemize}

\section{Face-StyleSpeech}

The primary focus of our method is to synthesize the speech from a face image for the target speaker, instead of the reference speech.
In this section, we begin by outlining a zero-shot TTS model~\cite{StyleSpeech, GradStyleSpeech}.
We then describe the face encoder, which is trained to generate a face latent vector.
To improve the face-to-speech mapping, we utilize the prosody encoder~\cite{megatts} to disentangle the speech style feature from the speech vector.
This disentanglement ensures an improved face-to-speech latent mapping by removing the need for encoding speaker-independent speech style features in the speech vector.
We illustrate our method in Figure~\ref{fig:method}.

\vspace{-0.1in}
\subsection{Preliminary: Zero-shot TTS Model}
\label{sec:tts}

We are given the training set $\{(\vx_i, \mY_i)\}_{i=1}^N$ including $N$ pairs of the text (phoneme) sequence $\vx \in \mathbb{R}^n$ and speech (mel-spectrogram) $\mY \in \mathbb{R}^{m \times b}$ where $b$ denotes the number of mel-filter banks (e.g., $b = 80$).
In the training phase, we train the TTS model to reconstruct the ground truth speech $\mY$ given the text $\vx$ and speech $\mY$ as inputs. Here, the zero-shot TTS model majorly consists of three modules: text encoder $F$, speech encoder $S$, and decoder $D$~\cite{StyleSpeech, GradStyleSpeech, AdaSpeech4}. 
The text encoder $F$ encodes the text sequence into representations $\mH_x = F(\vx;\theta_F) \in \mathbb{R}^{n \times d_F}$.
The speech encoder $S$ encodes the speech $\mY$ into the speech vector $\vs = S(\mY;\theta_S) \in \mathbb{R}^{d_S}$.
We utilize the \texttt{Aligner}~\cite{GradStyleSpeech, FastSpeech2}, which aligns the length of text and speech~\cite{Align}. 
The aligner outputs the duration of each phoneme: $\vd = [d_1, \ldots, d_n]$.
Then, we train the duration predictor to predict the duration of each phoneme given $\mH_x$ and $\vs$.
Given durations, we expand each text representation: $\bar{\mH}_x = \texttt{Expand}(\mH_x, \vd) \in \mathbb{R}^{m \times d_F}$.
The decoder $D$ reconstructs the target speech $\mY$ as follows: $\hat{\mY} = D (\bar{\mH}_x, \vs; \theta_{D})$.

In the inference stage, the model serves the text $\vx$ and the reference speech $\mY_r$ having different content to $\vx$.
The trained TTS model synthesizes the speech $\mY$ as follows:
\begin{equation}
    \mY = D \left(\texttt{Expand} \left( F \left( \vx;\theta_F^* \right), \hat{\vd} \right), S \left( \mY_r;\theta_S^* \right);\theta_D^* \right),
\end{equation}
where $\hat{\vd}$ is durations from the duration predictor, and $\theta^*$ denotes the trained parameter with the training set.

\subsection{Face Encoder Training for Face-to-Speech Mapping}

We now have a trained zero-shot TTS model capable of cloning the voice from a reference speech.
Then, our focus becomes determining how to generate the speech vector from a face image.
To this end, we train the face encoder $E$ by mapping the output from $E$ to the corresponding speech vector~\cite{Face2Speech, FR-PSS}.
We suppose that there exists a dataset $\{(\mY_i, \mI_i)\}_{i=1}^M$ having $M$ pairs of speech $\mY \in \mathbb{R}^{m \times b}$ and the face image $\mI \in \mathbb{R}^{3 \times 224 \times 224}$.
We embed the face image into the face latent vector using the face encoder $\vv = E(\mI; \theta_E)$ and the speech into the speech vector using the pre-trained speech encoder for the TTS model $\vs = S(\mY; \theta_S)$.
Then, we use the Mean Squared Error (\texttt{MSE}) loss and the negative cosine similarity (\texttt{cos}) loss for the face encoder training.
Furthermore, we utilize the contrastive learning objective inspired by CLIP~\cite{CLIP}.
The use of contrastive learning prevents the face vectors from collapsing into the same vector given the different faces, by forcing the face vector from different faces to have a different direction.
Combining all, we train the face encoder with the objective as follows:
\begin{align}
\label{eq:map}
   &\gL_{map} = \frac{1}{M} \sum_{i=1}^M \biggl((1 - \texttt{cos}(\vv_i, \vs_i)) + \texttt{MSE} (\vv_i, \vs_i) \\ \nonumber
   &-\log \frac{\exp(\texttt{cos}(\vv_i, \vs_i) / \tau)}{\exp(\texttt{cos}(\vv_i, \vs_i) / \tau) + \sum_{k=1}^K\exp(\texttt{cos}(\vv_i, \vs_k)/\tau)} \biggr), 
\end{align}
where $K$ is the number of negative samples and $\tau=0.07$.

Once trained, the face encoder $E^*$ with parameters $\theta_E^* =  \argmin_{\theta_E} \gL_{map}$ can generate the latent face vector $\vv$, which could replace the speech vector $\vs$ used in the TTS model described in Section~\ref{sec:tts}.
However, if the TTS model solely relies on the speech vector $\vs$ to encode both speech style and speaker identity, employing the face vector from $E^*$ could lead to unnatural speech synthesis as there is no strong correlation between the face image and speech style.

\vspace{-0.1in}
\subsection{Incorporating Prosody Feature for Improved Mapping}
\vspace{-0.05in}

We hypothesize that the issue mentioned in the previous section stems from the challenge of matching facial features with speech style features.
As a result, when mapping the face to the speech vector that has all prosodic features, the speech generated from an unseen face may exhibit abnormal prosody, as neither face nor text can comprehensively encode speech style.
Otherwise, the face encoder may only capture prominent facial attributes, such as gender and age, leading to the synthesis of speech that has similar voices for most face images.
As a solution, we propose the prosody encoder which aids the TTS model in encoding speech style independently of the speech vector derived from the speech encoder.

\subsubsection{Encoding Prosody with Vector Quantization}

During training, we can encode prosody related to speech style from the target speech $\mY$. 
Following Mega-TTS~\cite{megatts}, we use the prosody encoder which extracts the prosodic features from the speech into discrete codes.
Specifically, the prosody encoder $G$ encodes the speech $\mY$ into representations $\mH_y = G(\mY, \vd;\theta_G) \in \mathbb{R}^{n \times d_G}$, where we perform the phoneme-level pooling following the duration $\vd$.
This approach has its limitations, as the prosody encoder encodes both speaker-dependent and speaker-independent prosodic features.
To address this, we only use low-frequency features of mel-spectrogram $\mY$ and utilize the vector quantization \texttt{VQ}~\cite{VQ} as the information bottleneck for the prosody encoder as in previous works~\cite{ProsoSpeech, megatts}.
Through \texttt{VQ}, each prosody feature in $\mH_y$ is mapped to the code in the codebook $\mC = \{\vc_1, \vc_2, \ldots, \vc_T\}$. Therefore, we can encode $\mH_y$ as the discrete codes $\vh_y \in \mathbb{N}^n$ where $h \in [1, T] \  \forall h \in \vh_y$ with indices of mapped features. 
Then, we expand $\mH_y$ to fit the length of $\mY$: $\bar{\mH}_y = \texttt{Expand}(\mH_y, \vd)$. The decoder $D$ reconstructs $\mY$ with both prosody representations and the speech vector $\hat{\mY} = D (\bar{\mH}_x, \bar{\mH}_y, \vs; \theta_{D})$.

The use of prosody encoder $G$ enables the speech encoder $S$ to be able to encode only \textit{speaker-dependent features} into speech vector $\vs$, by disentangling the speaker-independent prosodic feature from the speech vector $\vs$.
As a result, during face encoder training in Eq.~\ref{eq:map}, the face vector $\vv$ can be better aligned with its speech vector $\vs$.

\subsubsection{Prosody Code Generation with Language Modeling}

In the inference stage, the trained TTS model synthesizes the speech from the target text $\vx$ and the face vector $\vv$ from the trained face encoder.
In this context, prosody codes for prosody modeling must be predicted from the target text $\vx$.
This is distinct from the training stage, where the prosody codes are extracted directly from the speech.
Therefore, we leverage the Prosody Language Model (PLM)~\cite{megatts} which generates the prosody codes $\vh_y$ for $\vx$ in an autoregressive manner as follows:
\begin{gather*}
    \vh_{y} = \texttt{PLM}(\vh^p_{y}, \mH^p_x, \mH_x;\theta_{PLM}), \quad \mH_x = F(\vx, \theta_F^*), \\
     \vh_y^p = \texttt{VQ}(G(\mY^p, \hat{\vd}; \theta_G^*), \quad \mH_x^p = F(\vx^p, \theta_F^*), 
\end{gather*}
where $\hat{\vd}$ is predicted durations of $\vx^p$ from the duration predictor, $\mY^p$ is the prompt speech and its transcript $\vx^p$.
Notably, the \textbf{prompt speech $\mY^p$ can be any speech} since it is only used to generate prosody codes, not the speaker-related features.
Therefore, we can use randomly sampled speech from the training dataset as a prompt speech.
For PLM training, we use the teacher forcing on the TTS training dataset~\cite{megatts}.
After generating the prosody codes using PLM, we decode $\vh_y^p$ into representations $\mH_y$ with the codebook used in \texttt{VQ}.
We then generate natural speech with a particular voice from an image:
\begin{equation}
    \mY = D\left(\bar{\mH}_x, \bar{\mH}_y, E\left(\mI;\theta_E^*\right);\theta_D^*\right),
\end{equation}
where $\bar{\mH} = \texttt{Expand}(\mH, \hat{\vd})$ with predicted durations $\hat{\vd}$.

\section{Experiment}

\subsection{Experimental Setup}

\textbf{Implementation Details.}
We use the Grad-StyleSpeech~\cite{GradStyleSpeech} for the TTS model and follow the training setups in the original paper.
For the face encoder, we use the SyncNet~\cite{SyncNet} following the Face-TTS~\cite{ImaginaryVoice}.
We train the face encoder for 3k steps with batch size 256, Adam optimizer, and a learning rate of $10^{-4}$.
For the prosody encoder and language model, we follow the implementation details in Mega-TTS~\cite{megatts}.
We use the same audio processing setup with a sampling rate of 16khz following previous works~\cite{StyleSpeech, GradStyleSpeech}.
We use the HiFi-GAN~\cite{HifiGAN} as a vocoder.

\textbf{Dataset.}
We train the TTS model on the English speech dataset LibriTTS-R~\cite{LibriTTS-R}, which contains 245 hours of audiobook recordings from 553 speakers. 
We use the \texttt{clean-100} and \texttt{clean-360} subsets for the training.
Then, we use the audio-visual dataset VoxCeleb~\cite{VoxCeleb} which includes the interview clips of 1,251 celebrities for the training of the face encoder.
For evaluation, we use 20 sampled faces from the VoxCeleb 2~\cite{VoxCeleb2} which are not used in the training.

\textbf{Baselines.} 
\textbf{(1) Face-TTS~\cite{ImaginaryVoice}.} Face-styled TTS model based on the Grad-TTS~\cite{GradTTS} where the face encoder is jointly trained with the TTS model on the LRS3~\cite{LRS3} dataset.
\textbf{(2) Grad-StyleSpeech.} Vanilla Grad-StyleSpeech~\cite{GradStyleSpeech} model without any additional module, but inputs the face vector from the face encoder as a condition.
\textbf{(3) Grad-StyleSpeech + CLAP.} Same as (2), but we use the text encoder from CLAP-Speech~\cite{CLAPSpeech} in the TTS model to consider the prosody variants from the text content.
\textbf{(4) Face-StyleSpeech.} Our proposed model with the prosody encoder and Prosody Language Model~\cite{megatts} in the Grad-StyleSpeech to consider the prosody.

\textbf{Evaluation Metric.}
We evaluate face-conditioned TTS models using the following metrics.
(1) Character Error Rate (\textbf{CER}): Measures intelligibility using whisper~\cite{whisper} ASR model to transcribe synthesized speech.
(2) Mean Opinion Score (\textbf{MOS}): Assesses subjective naturalness through human evaluation on a 5-point scale.
(3) Speaker Embedding Cosine Similarity (\textbf{SECS}): Evaluates voice similarity between synthesized and reference speech by measuring cosine similarity of speaker embeddings using Resemblyzer\footnote{\href{https://github.com/resemble-ai/Resemblyzer}{https://github.com/resemble-ai/Resemblyzer}}.

Face-conditioned TTS models should not generate identical voices for different faces, as each person has a unique voice identity. To assess this, we use: (4) Speaker Embedding Diversity (\textbf{SED}): Measures pairwise similarity of synthesized voices from different face images; lower SED indicates greater voice diversity. Additionally, models should generate speech that matches the face image. We test this with: (5) Matching Preference Test: Evaluators choose between synthesized speech or face images that best match each other, based on 5 samples with 10 evaluators. Evaluations are conducted on gender-matched face-voice pairs to avoid trivial cases.

\begin{table}
	\centering
        \setlength{\tabcolsep}{1.0em}
	\small
        \caption{\textbf{Evaluation on Unseen Faces.} Experimental results on unseen face images from VoxCeleb 2~\cite{VoxCeleb2} where the models are not trained. Note that Face-TTS$^\dagger$ is trained on the LRS3~\cite{LRS3} dataset.}
        \vspace{-0.1in}
	\resizebox{0.50\textwidth}{!}{
	\begin{tabular}{lcccc}
		\toprule
 {\textbf{Model}}  & MOS ($\uparrow$) & CER ($\downarrow$) & SECS ($\uparrow$) & SED ($\downarrow$) \\
		\midrule[0.4pt]
            \textbf{Face-TTS}$^\dagger$ & 2.45 $\pm$ 0.17 & 4.47 & 70.74 & 82.57 \\
		\textbf{Grad-StyleSpeech (GSS)}  & 2.65 $\pm$ 0.18 & 7.59 & 72.25 & 89.94 \\
		\textbf{GSS + CLAPSpeech} & 2.52 $\pm$ 0.16 & 5.70 & 71.56 & 91.71 \\
            \textbf{Face-StyleSpeech} \textit{(ours)} & \bf 3.72 $\pm$ 0.16 & \bf 1.39 & \bf 74.14 & \bf 80.45 \\
		\bottomrule
	\end{tabular}
	}
 	\vspace{-0.15in}
 	\label{tab:main}
\end{table}

\begin{figure}
    \centering
    \includegraphics[width=1.0\linewidth]{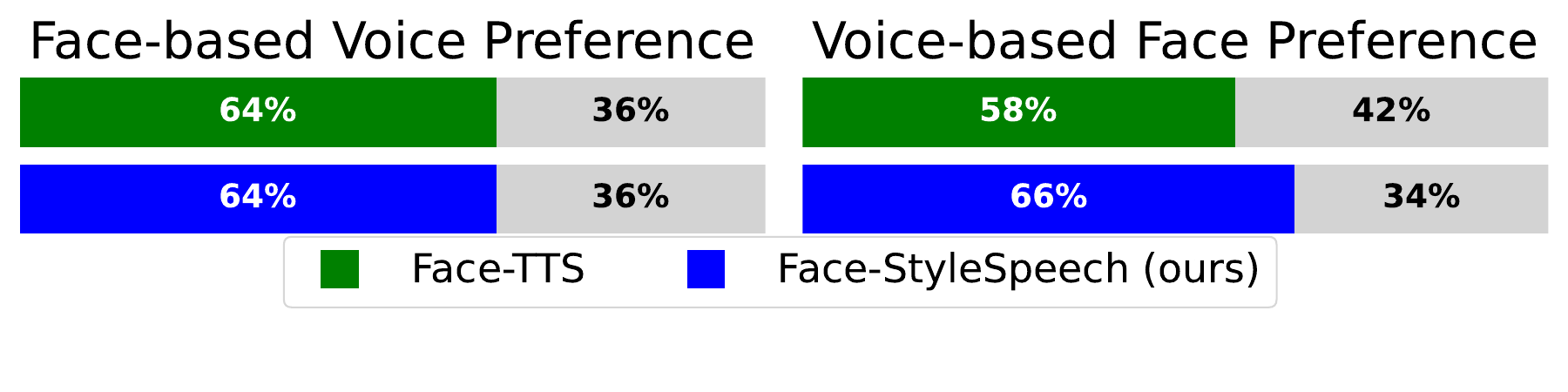}
    \vspace{-0.4in}
    \caption{\textbf{Preference Test.} Results of (Left) Face-based Voice Preference 
and (Right) Voice-based Face Preference Tests.
    }
    \vspace{-0.2in}
    \label{fig:matching}
\end{figure}

\subsection{Experimental Results}
\textbf{Main Experiments} 
Table~\ref{tab:main} shows our experimental results for text-to-speech synthesis conditioned on face images, particularly those on which the models have not been trained.
As shown in Table~\ref{tab:main}, our Face-StyleSpeech shows a higher MOS and a lower CER than baselines, indicating that the speech synthesized by Face-StyleSpeech sounds more natural and intelligible than baselines. 
Furthermore, Face-StyleSpeech outperforms baselines in terms of speaker similarity, with higher SECS indicating that the synthesized speech is more similar to the target speaker's real voice.

\textbf{Matching Preference}
In Figure~\ref{fig:matching}, we plot the results of matching preference tests. 
In both Face-TTS and Face-StyleSpeech, evaluators select the correct answer with 64\% accuracy in the face-based voice preference test, even if we only use pairs having the same gender for more difficult cases.
In the voice-based face preference test, however, evaluators select 8\% more correct answers when the speech is generated by Face-StyleSpeech rather than Face-TTS, demonstrating that our Face-StyleSpeech synthesizes the speech with a voice more matched to the given face.
Lastly, as shown in the SED metric of Table~\ref{tab:main}, our Face-StyleSpeech generates more diverse voices than baselines.
Experimental results support our hypothesis that disentangling prosody from the speech vector improves face-to-voice mapping, resulting in the TTS model that generates more natural, diverse, and matched speech from a face image.
Please refer to the demo page\footnote{\href{https://face-stylespeech.github.io}{https://face-stylespeech.github.io}} for samples.

\begin{figure}
    \centering
        \includegraphics[width=0.85\linewidth]{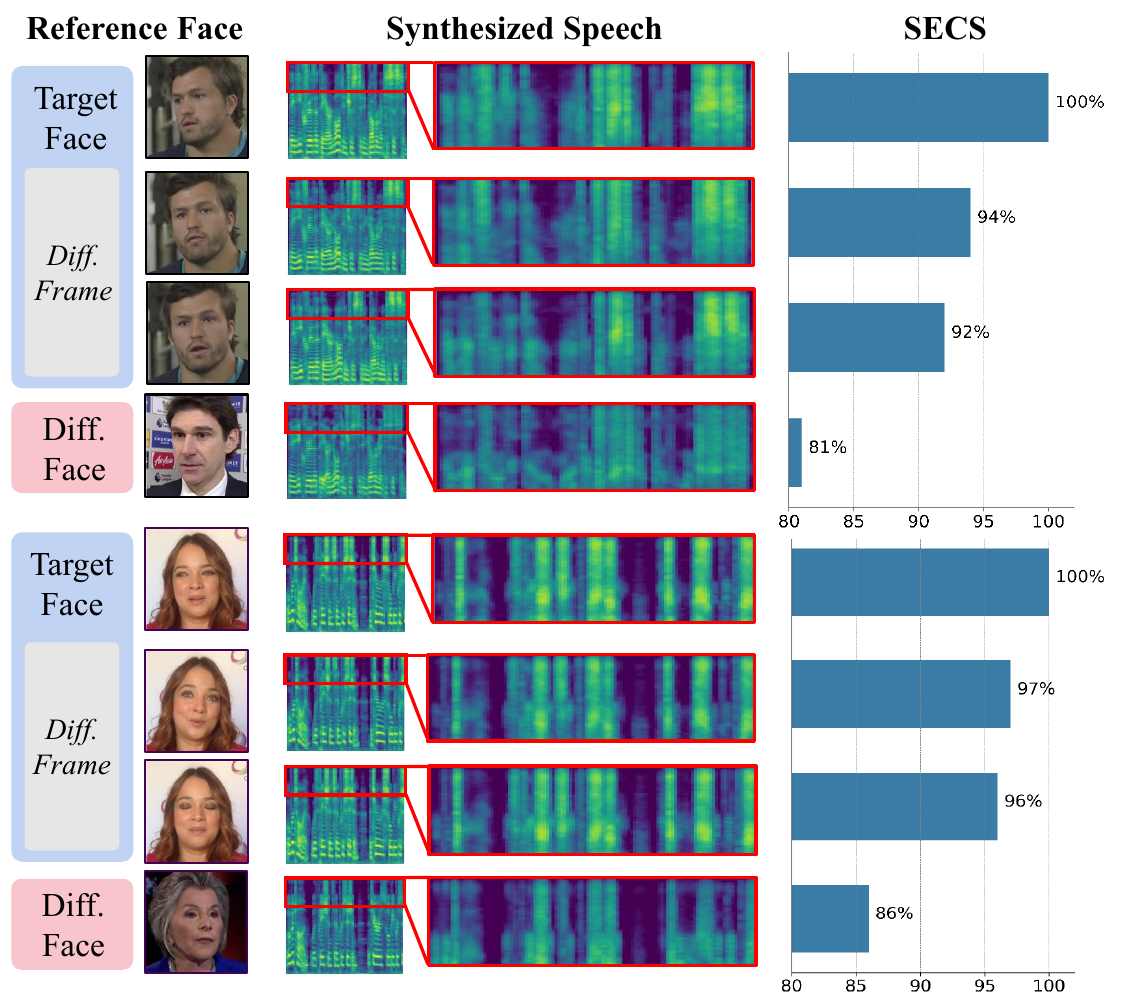}
    \vspace{-0.12in}
    \caption{\textbf{Analysis on Effects of Different Frames.} We visualize the mel-spectrograms of synthesized speech from the face images. On the right, we plot the SECS between speech samples and the speech from the target face.}
    \vspace{-0.1in}
    \label{fig:diff_frame_same_voice}
\end{figure}

\textbf{Consistency Test}
In face-conditioned TTS models, the same voice must be synthesized from images of the same person's face as much as different voices must come from images of different people's faces. For instance, video datasets like VoxCeleb~\cite{VoxCeleb, VoxCeleb2} contain multiple frames of the same individual. It is essential that, regardless of the frame chosen as input, a consistent voice is produced. We conducted experiments to demonstrate that our model possesses this capability, as shown in Figure~\ref{fig:diff_frame_same_voice}. 
We sample three different frames each for males and females and synthesize speech from them.
The resulting mel-spectrograms show similar trends, indicating that Face-StyleSpeech can maintain voice consistency across different frames of the same person.

\textbf{Ablation Study} In Table~\ref{tab:ablation}, we outline the results of an ablation study for the loss functions in Equation~\ref{eq:map}, which is used in the face encoder training. 
We utilize Face-StyleSpeech in this ablation study. In addition to contrastive learning loss, we also experiment with triplet loss~\cite{FR-PSS}.
Our findings indicate that contrastive learning is important, especially when it comes to voice diversity.

\begin{table}
	\centering
	\small
        \caption{\textbf{Ablation Study.} Ablation study results on the loss functions used in the face encoder training in Equation~\ref{eq:map}.}
        \vspace{-0.1in}
	\resizebox{0.95\linewidth}{!}{
	\begin{tabular}{lccc}
		\toprule
            $\gL_{map}$    & CER ($\downarrow$) & SECS ($\uparrow$) & SED ($\downarrow$) \\
		\midrule[0.4pt]
            $\texttt{MSE}$ + neg. Cosine & 1.98 & 72.11  & 90.64 \\
		\quad + Triplet Loss & 2.39  & 72.04  & 87.49  \\
		\quad + Contrastive Loss & \bf 1.39  & \bf 74.14  & \bf 80.45  \\
		\bottomrule
	\end{tabular}
	}
 	\vspace{-0.17in}
 	\label{tab:ablation}
\end{table}

\section{Conclusion}
In this work, we introduced Face-StyleSpeech, a zero-shot TTS model that synthesizes natural speech from face images instead of reference speech.
The key challenge we addressed was the weak correlation between face images and entire prosody features encapsulated within the speech vector. 
To overcome this, we introduced a prosody encoder in the TTS model to separate speech style features from the speech vector, thereby enhancing the model's ability to generate more natural speech by improving the face-to-speech mapping. 
Our experimental results demonstrate that Face-StyleSpeech is effective in generating natural-sounding speech that captures the imagined voice of provided faces well, even those not seen during training. 

\newpage
\bibliographystyle{IEEEtran}
\bibliography{reference}

\end{document}